\documentclass{article}
\usepackage{spconf,amsmath,graphicx,xcolor,algorithm2e,algorithmic}
\usepackage{hyperref}

\renewcommand{\normalsize}{\fontsize{9}{11}\selectfont}

\title{An adversarial learning framework for preserving users' anonymity in face-based emotion recognition}

\name{Vansh Narula, Zhangyang (Atlas) Wang, and Theodora Chaspari\thanks{Thanks to the Texas A\&M University Program to Enhance Scholarly and Creative Activities (PESCA) for supporting this research.}}
\address{Computer Science \& Engineering, Texas A\&M University}
\begin{document}
%
\maketitle
\begin{abstract}
\vspace{-5pt}
Image and video-capturing technologies have permeated our every-day life. Such technologies can continuously monitor individuals' expressions in real-life settings, affording us new insights into their emotional states and transitions, thus paving the way to novel well-being and healthcare applications. Yet, due to the strong privacy concerns, the use of such technologies is met with strong skepticism, since current face-based emotion recognition systems relying on deep learning techniques tend to preserve substantial information related to the identity of the user, apart from the emotion-specific information. This paper proposes an adversarial learning framework which relies on a convolutional neural network (CNN) architecture trained through an iterative procedure for minimizing identity-specific information and maximizing emotion-dependent information. The proposed approach is evaluated through emotion classification and face identification metrics, and is compared against two CNNs, one trained solely for emotion recognition and the other trained solely for face identification. Experiments are performed using the Yale Face Dataset and Japanese Female Facial Expression Database. Results indicate that the proposed approach can learn a convolutional transformation for preserving emotion recognition accuracy and degrading face identity recognition, providing a foundation toward privacy-aware emotion recognition technologies.
\end{abstract}

\vspace{-5pt}
\section{Introduction}
\label{sec:intro}
\vspace{-8pt}
Image and video-capturing devices have become increasingly ubiquitous and pervasive. From the millions of surveillance cameras installed all over the world to the newly introduced smart home devices, such ambulatory recording technologies allow the continuous monitoring of individuals over long periods of time. Beyond well-established applications related to security monitoring and community safety, the continuous capturing of human expression in real-life environments can promote healthcare and well-being applications~\cite{leon2011prospect,preschl2011health}. For example, the monitoring of facial expressions and body gestures in a continuous manner can capture momentary and longitudinal patterns of human emotion, which can be reflective of users' stress, depression, or even suicidal risk, therefore rendering such information a valuable biomarker for predicting and potentially intervening upon individuals’ mental and emotional health~\cite{narayanan2013behavioral}.

Despite the premises, the barrier of confidentiality and anonymity inherent in these smart-monitoring applications is an issue with various social and cultural implications preventing their wide adoption. Users are often skeptical of such technologies, since they are afraid that facial information relevant to their identity will be permanently stored in third-party servers or will be abused by hacker attacks~\cite{KshetriV18a}. This does not come as a surprise, since state-of-the art computer vision systems for emotion recognition relying on rich facial features, such as the histograms of oriented gradients (HoG)~\cite{dahmane2011emotion}, and representation learning models, such as convolutional neural networks (CNN)~\cite{nguyen2017deep}, tend to preserve a significant amount of facial information related to the identity of the user. This privacy compromising landscape renders essential the design of novel machine learning systems that conceal one's identity, while at the same time preserve useful information for emotion recognition.

Extensive work on human activity recognition has leveraged CNN-based architectures to learn image transformations for a given classification or regression task~\cite{C18}. In fact, previous work suggests that the convolutional transformations of the CNN are able to capture general and highly reusable information~\cite{gatys2016image}, thus the CNN layers can be pre-trained on one task and subsequently fine-tuned to another proximal one~\cite{gu2018recent}. Although this is a highly desirable property for many applications, the ability of CNNs to retain reusable information can pose an innate threat in cases with high sensitive data, since the learning of the convolutional transformation in terms of emotion-specific information might also preserve information relevant to an individual's identity. To overcome this challenge, prior work has used pre-defined image degradation transformations in order to reduce the total amount of information preserved in an image, as well as an optimization framework to learn an image transformation from a set of data for the corresponding tasks of interest~\cite{Wu_2018_ECCV,C12}. Despite the promising results, this work has solely focused on the task of human activity identification, which involves the presence of multiple individuals in a frame recorded from a long distance. In contrast, the problem of privacy-preserving emotion recognition presents an additional set of unique challenges, as it depends on learning subtle facial emotional expressions and usually relies on data from cameras placed at a close proximity to one's face.

In order to balance the trade-off between user anonymity and emotion-specific face characteristics, we propose an adversarial learning algorithm that learns an image transformation to degrade sensitive information relevant to the user identity and preserve emotion-dependent information. A hybrid neural network architecture is comprised of convolutional layers $f_{W_c}$, followed by two parts of fully-connected layers, one for emotion classification $f_{W_e}$ and another for face identity recognition $f_{W_i}$. The convolutional layers $f_{W_c}$ are learned so that they can degrade face identity-dependent information for any possible user-dependent transformation $f_{W_i}$ and at the same time preserve emotion-dependent information in $f_{W_e}$. Our results in two datasets indicate that the proposed approach can achieve emotion recognition performance equivalent to the one of a CNN fully trained on emotion recognition and at the same time, significant degradation in face identification, indicating the feasibility of the proposed framework for promoting user privacy in visual applications.

\vspace{-7pt}
\section{Relation to prior Work}
\label{sec:priorwork}
\vspace{-8pt}
The automatic recognition of facial expressions has always been an interesting problem in computer vision. Prior work has tackled this problem by engineering appropriate features, such as Histogram of Oriented Gradients (HOG) and eigen-faces~\cite{C17}. Other approaches have leveraged the frequency characteristics of an image through Gabor filters and Wavelets~\cite{C16}. Finally, more recent techniques focus on the automatic learning of features through CNN-based architectures, such as the Resnet, MobileNet and Inception Network~\cite{C18}. These approaches tend to capture facial features that are considered important for both emotion recognition and face identification.

In the light of these, prior work has approached the problem of privacy preservation in general human activity recognition from two different approaches, one relying on pre-defined transformations of an image, while the other leveraging supervised optimization approaches. Image transformation approaches have attempted to increase the amount of uncertainty throughout the image by adding noise~\cite{C4} or performing filtering operations~\cite{C1}. They have further tried to decrease the resolution of the facial area of a person~\cite{C12}, as well as to encode the change in successive images as the input to the system, compared to the image pixels themselves~\cite{Steil_2019}. Supervised learning approaches have formulated this as an optimization problem, leveraging an adversarial learning framework for learning appropriate degradations of images to increase a target utility metric and minimize a privacy-based metric~\cite{C5,Wu_2018_ECCV,hamm2017minimax,raval2017protecting}.
.

The proposed work advances existing literature in the following ways: (1) In contrast to previous work on privacy preservation for general human activity recognition, this paper proposes a privacy-preservation system specifically for the task of emotion recognition. This task is highly dependent on subtle facial characteristics, for which it is much more difficult to learn appropriate degradation transformations; (2) While previous work has focused on data obtained with surveillance cameras or distant cameras capturing the entire body from one or multiple users~\cite{Wu_2018_ECCV,hamm2017minimax,raval2017protecting}, this paper relies on cameras placed in close proximity to a user's face, therefore preserving a high amount of identity-specific information; (3) Most of the work does not to provide a clear method of evaluating the trade-off between the degradation of utility-based information and preservation of user identity~\cite{C1,C4}. The proposed adversarial learning framework results in a convolutional transformation that attempts to degrade user-specific information for any of the subsequent fully-connected layers. The output of the convolution is fed into two classifiers, one for emotion recognition and the other for face identification. These accuracies can quantify the amount of identity- and emotion-specific information preserved in the CNN.

\vspace{-7pt}
\section{Methodology}
\label{sec:majhead}
\vspace{-8pt}

\subsection{Quantifying the identity-specific information of an emotion-specific convolutional base}
\label{subsec:subhead1}
\begin{figure*}[t]
\vspace{-10pt}
\begin{minipage}[t]{0.49\linewidth}
  \centering
  \centerline{\includegraphics[trim = 0mm 0mm 0mm 0mm, clip=true, width=0.98\linewidth]{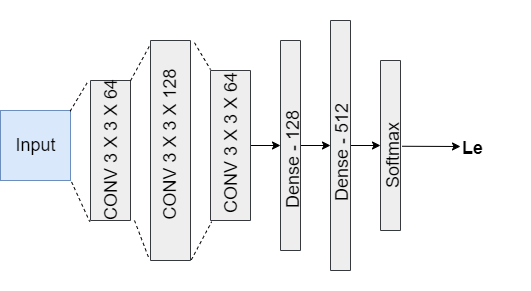}}
  \centerline{\small{(a) {\it Emotion} model}}\medskip
\end{minipage}
\begin{minipage}[t]{0.49\linewidth}
  \centering
  \centerline{\includegraphics[trim = 0mm 0mm 0mm 0mm, clip=true, width=0.98\linewidth]{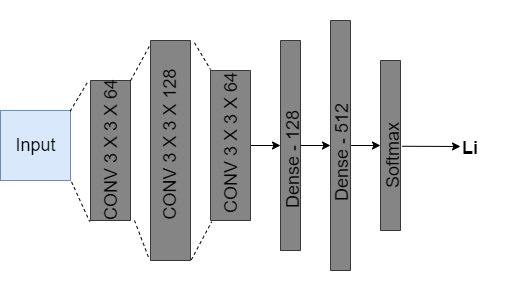}}
  \centerline{\small{(b) {\it Face} model}}\medskip
\end{minipage}
\begin{minipage}[t]{0.49\linewidth}
  \centering
  \centerline{\includegraphics[trim = 0mm 0mm 0mm 0mm, clip=true, width=0.98\linewidth]{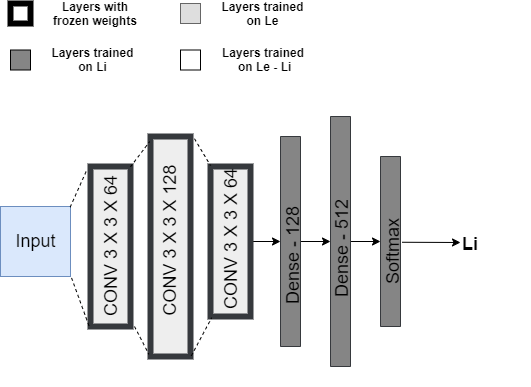}}
  \centerline{\small{(c) {\it Emotion2Face} model}}\medskip
\end{minipage}
\begin{minipage}[t]{0.49\linewidth}
  \centering
  \centerline{\includegraphics[trim = 0mm 0mm 0mm 0mm, clip=true, width=1.1\linewidth]{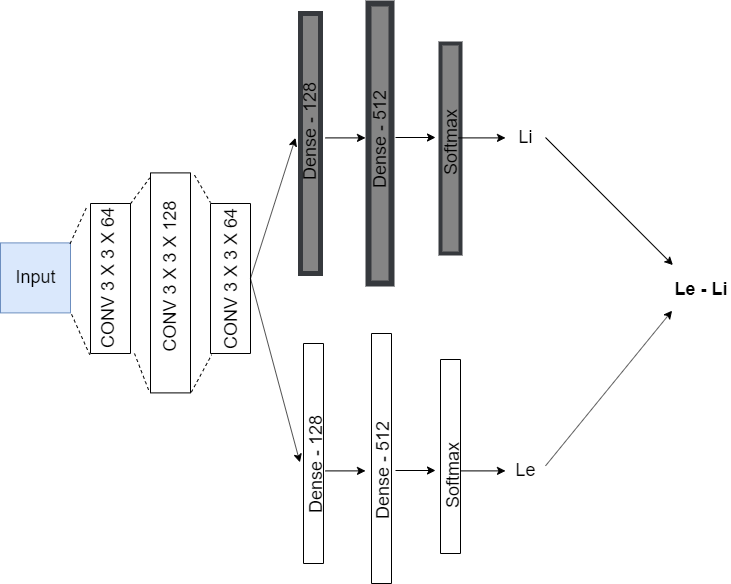}}
  \centerline{\small{(d) {\it Hybrid} model}}\medskip
\end{minipage}
\vspace{-10pt}
\caption{Schematic representation of the: (a) {\it Emotion} model, trained on emotion recognition; (b) {\it Face} model, trained on face identification; (c) {\it Emotion2Face} model, trained on emotion and fine-tuned on face identification; and (d) {\it Hybrid} model, trained on an iterative adversarial framework for user identity-preserving emotion recognition.}
\label{fig:Architectures}
\vspace{-7pt}
\end{figure*}

\vspace{-8pt}
We first train two separate CNNs, one for emotion classification, referred to as ``{\it Emotion}" (Fig.~\ref{fig:Architectures}a), and another for face identity recognition, referred to as ``{\it Face}" (Fig.~\ref{fig:Architectures}b). Each of the CNNs includes a set of convolutional layers, followed by fully-connected ones. The {\it Face} model serves as a baseline to quantify the amount of information specific to the user identity that can be captured through the convolutional layers fully trained for face identification, as measured by the corresponding face identification accuracy. Similarly, the emotion classification accuracy of the {\it Emotion} model can approximate the degree of emotion-specific information present in the convolutional layers of a CNN fully trained on emotion. These accuracies will serve as baselines, against which the proposed adversarial framework (Section~\ref{subsec:subhead2}) will be compared for its ability to preserve the emotional information and eliminate user-specific information.

In order to quantify the amount of identity-specific information built-in in the emotion recognition model {\it Emotion}, we keep its convolutional layers frozen and fine-tune its fully-connected layers for the task of face identification. This model, which will be referred to as ``{\it Emotion2Face}" (Fig.~\ref{fig:Architectures}c), can capture the amount of identity-specific information present in the convolutional base of the CNN trained for emotion recognition. The face identification accuracy of the {\it Emotion2Face} model will be employed as an approximate measure for that. High values of this measure suggest that the convolutional base of the emotion-specific CNN retains a large degree of face identity information, while low values indicate that the convolutional base does not preserve much user-specific information.

\vspace{-10pt}
\begin{algorithm}[t]
\begin{algorithmic}[1]
\REQUIRE Image $\mathbf{x}$, emotion label $y_e$, user label $y_i$, hyperparameters $\alpha$, $\beta$, $T$
\STATE Initialize $W_c$ (convolutional weights), $W_e$ (emotion classification weights), $W_i$ (face identificaiton weights)  with multi-task learning
\resizebox{0.91\hsize}{!}{$\min_{\{W_c,W_e,W_i\}}\{L_e\left(f_{W_e}(f_{W_c}(\mathbf{x})), y_e\right) + \alpha L_i\left(f_{W_i}(f_{W_c}(\mathbf{x})), y_i\right)\}$}
\FOR{$t=1,\ldots,T$}
  \STATE Freeze $W_i$
  \STATE Learn $W_c$ and $W_e$ using adversarial learning
  \resizebox{0.91\hsize}{!}{$\min_{\{W_c,W_e\}}\{L_e\left(f_{W_e}(f_{W_c}(\mathbf{x})), y_e\right) - \beta L_i\left(f_{W_i}(f_{W_c}(\mathbf{x})), y_i\right)\}$}
  \STATE Freeze $W_c$ and $W_e$
  \STATE Learn $W_i$
$\min_{W_i}\{L_i\left(f_{W_i}(f_{W_c}(\mathbf{x})), y_i\right)\}$
\ENDFOR
\end{algorithmic}
\caption{Adversarial learning for anonymity preserving emotion recognition\label{Alg:1}}
\end{algorithm}

\vspace{-3pt}
\subsection{User anonymity preserving emotion classification}\vspace{-7pt}
\label{subsec:subhead2}
We propose an adversarial framework which learns a transformation to maximize the emotion-specific information and minimize the identity-specific information related to the privacy-preservation task (Algorithm~\ref{Alg:1}). We will use a hybrid architecture including a shared convolutional base, connected to two different sets of fully connected layers, one for emotion classification and one for face recognition, which will be referred to as ``{\it Hybrid}" (Fig.~\ref{fig:Architectures}d). The proposed adversarial approach contains two key components: (1) We learn appropriate transformations both at the convolutional and fully-connected layers. Since the convolutional layers preserve a large portion of the information in an image~\cite{gatys2016image}, they also tend to be more prone in capturing the user identity. If we apply the adversarial learning solely on the fully-connected layers, then the corresponding weights will quickly reach to zero, while the identity-dependent information will be preserved in the convolutional part; (2) We further aim to learn a convolutional base that is able to eliminate identity-specific information when followed by any fully-connected layer. The intuition behind this lies in that the fully-connected layers of a reliable face recognition model will be able to extract information specific to the user identity from a given convolutional base. For this reason, we employ an iterative procedure to learn a convolutional transformation, based on which there is no fully connected layer able to extract face identity-specific information.

In the following, let $\mathbf{x}$ be an input image with face identity label $y_i$ and emotion label $y_e$. Also let $f_{W_c}(\mathbf{x})$ be the transformation of the shared convolutional layers, represented by the weights $W_c$. The output of the convolutional layers is fed to two fully connected layers, $f_{W_e}(f_{W_c}(\mathbf{x}))$ and $f_{W_i}(f_{W_c}(\mathbf{x}))$, performing the emotion-specific and face identity-specific transformations using weights $W_e$ and $ W_i$. We further assume that $L_e$ and $L_i$ are the cross-entropy loss of emotion classification and face identity recognition. We initialize the weights $W_c$, $W_e$, and $W_i$ by jointly training the hybrid architecture as a multi-task learning task:\vspace{-5pt}
\begin{equation}\begin{aligned}
\resizebox{0.91\hsize}{!}{$\min_{\{W_c,W_e,W_i\}}\{L_e\left(f_{W_e}(f_{W_c}(\mathbf{x})), y_e\right) + \alpha L_i\left(f_{W_i}(f_{W_c}(\mathbf{x})), y_i\right)\}$}
\end{aligned}\label{eq:1}\end{equation}
where $\alpha$ is the hyper-parameter that balances between positive emotion loss and positive face identity loss. We then freeze the weights $W_i$ of fully-connected layers for identity recognition and perform an adversarial learning process to optimize the following criterion:\vspace{-5pt}
\begin{equation}\begin{aligned}
\resizebox{0.91\hsize}{!}{$\min_{\{W_c,W_e\}}\{L_e\left(f_{W_e}(f_{W_c}(\mathbf{x})), y_e\right) - \beta L_i\left(f_{W_i}(f_{W_c}(\mathbf{x})), y_i\right)\}$}
\end{aligned}\label{eq:2}\end{equation}
where $\beta$ balances between positive emotion loss and negative face identity loss. This allows to initiate the adversarial training with a reliable set of weights $W_i$ for face identification. At the same time, this prevents $W_i$ from becoming zero, which would result in an ``artificially" successful adversarial learning with the face identity information being likely to remain in the convolutional base.

The above adversarial learning ensures that the model is trained in way that it conceals face identity-specific information from the current transformation $f_{W_i}$, but we would like the model to generalize and hence it should able to ``fool" any transformation, i.e. any possible value of $W_i$ should not be able to extract any face specific information from the convolutional base. In order to achieve that, we would ideally have to perform the adversarial training for all possible values of $W_i$, which is computationally not feasible. For this reason, we freeze the convolutional base $W_c$ and emotion-specific fully-connected layers $W_e$, and learn the face identity weights $W_i$, such that
$\min_{W_i}\{L_i\left(f_{W_i}(f_{W_c}(\mathbf{x})), y_i\right)\}$.
After this we have another reliable estimate of $W_i$ for face identity recognition, and thus can restart the adversarial training to further learn the weights $W_c$ and $W_e$. This process is repeated $T$ times until the transformation $f_{W_i}$ is no longer able to achieve good face recognition accuracy, suggesting that the face identity-specific information is either lost or hard to be recovered.

Given the final learned convolutional transformation $f_{W^*_c}(\mathbf{x})$, we evaluate its ability to preserve emotion-specific information and eliminate information related to the identity of a speaker. For this reason, we add two different sets of feedforward layers $f_{W'_e}$ and $f_{W'_i}$, whose weights $W'_e$ and $W'_i$ are learned for emotion recognition (i.e., $\min_{W'_e}{L_e(f_{W'_e}(f_{W^*_c}(\mathbf{x})), y_e)}$) and face identification (i.e., $\min_{W'_i}{L_i(f_{W'_i}(f_{W^*_c}(\mathbf{x})), y_e)}$), respectively. These models will be referred to as ``{\it Hybrid2Emotion}" and ``{\it Hybrid2Face}" and their final accuracies will be reported.

\vspace{-7pt}
\section{Experiments}
\label{sec:experiments}

\vspace{-7pt}
\subsection{Data description and pre-processing}
\vspace{-8pt}
\label{subsec:DataDescr}
We tested our approach on the Japanese Female Facial Expression (JAFFE) database~\cite{lyons1998japanese} and the Yale Face Dataset (YALE)~\cite{C7}. JAFFE included 10 female users and 7 emotions (neutral, sadness, surprise, happiness, fear, anger, and disgust), with a total of 213 static images. For YALE, we used only the images which include labels for both the user and emotion categories. These include 15 male and female users and 4 emotion classes (happy, sad, normal, surprised), resulting in a total of 60 images.  Since the number of images in both datasets was small for a deep learning model to be adequately trained, we used data augmentation techniques related to random rotation, horizontal flip, and random noise addition to generate a total of 3038 and 3033 images for JAFFE and YALE, respectively~\cite{Shorten2019}.

\vspace{-7pt}
\subsection{Experimental setting}
\vspace{-8pt}
\label{subsec:results}
For all experiments we used a 10-fold cross-validation, with the same set of train and test images per fold across all systems and making sure that no samples generated from the same original image after data augmentation are present in the test set concurrently. The problem formulation does not qualify for a leave-one-subject-out cross-validation, since we need the convolutional transformation to learn from every subject in order to be able to calculate the corresponding face-recognition accuracy. Similarly we need the model to be trained for every emotion. All the considered models (i.e., {\it Emotion}, {\it Face}, {\it Emotion2Face}, and {\it Hybrid}) included 3 convolutional followed by 3 fully-connected layers. For the case of the {\it Hybrid} model, two streams of fully connected layers were included, one for emotion and the other for user classification. The activation function of all the hidden layers was ReLU, while the output layer had a softmax activation. The filter of the convolution operation was equal to 3 with a stride length of 1. The number of nodes for each layer is depicted in Fig.~\ref{fig:Architectures}. The hyper-parameters balancing the {\it Hybrid} model's ability to learn between the emotion and the user categories in (\ref{eq:1}) and (\ref{eq:2}) were set to to $\alpha=0.5$ and $\beta=1$, respectively. The number of iterations for the adversarial learning optimization was 50 and 20 for the JAFFE and YALE, respectively.

\vspace{-7pt}
\subsection{Results}
\vspace{-8pt}
\label{subsec:results}
Our results for the JAFFE and YALE datasets are presented in Tables~\ref{tab: table1} and~\ref{tab: table2}. All results reflect simple classification accuracies, since the distribution of samples for the user and emotion categories was balanced for both datasets. Face identity recognition appears to be an easy task for both datasets, yielding classification accuracies of 99\% and 87\%. Emotion classification depicts higher accuracy for JAFFE compared to YALE, potentially due to the high variability of the latter. When the {\it Emotion} model is fine-tuned for face identity recognition, the corresponding accuracies of the {\it Emotion2Face} model (90\% for JAFFE, 78\% for YALE) suggest that there is a substantial amount of user-dependent information in the convolutional layer of the {\it Emotion} model reflecting a user's identity. The proposed adversarial learning framework on the JAFFE dataset yields low face identity recognition (i.e., 25\% for {\it Hybrid2Face}) and emotion recognition comparable to the performance of the original {\it Emotion} model (i.e., 84\% for {\it Emotion}, 81\% for {\it Hybrid2Emotion}). Similarly, for the YALE dataset, the adversarial learning framework was able to preserve the emotion recognition performance (i.e., 44\% for {\it Emotion}, 49\% for {\it Hybrid2Emotion}) and degrade the face recognition (i.e., 1\% for {\it Hybrid2Face}).

\begin{table}[]
\scriptsize
\caption {Emotion and user classification accuracies in JAFFE} \label{tab: table2}
\begin{center}
\begin{tabular}{|p{0.10\linewidth}
|p{0.10\linewidth}
|p{0.15\linewidth}
|p{0.18\linewidth}
|p{0.15\linewidth}|}
\hline
{\it Face} & {\it Emotion} & {\it Emotion2Face} & {\it Hybrid2Emotion} & {\it Hybrid2Face} \\ \hline
99.26 & 84.27 & 90.41 & 81.57 & 25.79                        \\ \hline
\end{tabular}
\end{center}
\vspace{-20pt}
\end{table}

\begin{table}[]\vspace{-3pt}
\scriptsize
\caption {Emotion and user classification accuracies in YALE} \label{tab: table1}
\begin{center}
\begin{tabular}{|p{0.10\linewidth}
|p{0.10\linewidth}
|p{0.15\linewidth}
|p{0.18\linewidth}
|p{0.15\linewidth}|}
\hline
{\it Face} & {\it Emotion} & {\it Emotion2Face} & {\it Hybrid2Emotion} & {\it Hybrid2Face} \\ \hline
87.96 & 44.611 & 78.44 & 49.122 & 1.00                        \\ \hline
\end{tabular}
\end{center}
\vspace{-20pt}
\end{table}

\vspace{-7pt}
\section{Discussion}
\vspace{-8pt}
Although the proposed framework provides encouraging results, the current study presents the following limitations. The YALE and JAFFE datasets are collected in laboratory conditions and include acted emotions. It would be beneficial to test the proposed approach in real-life applications with spontaneous emotional expressions and a larger number of samples. Surprisingly, we were not able to find many image datasets which contain labels of user identity and emotion for the same image, comprising an impediment for this type of work. Preserving information related to the user identity when designing emotional datasets would provide beneficial benchmarks for this task. Another limitation of this study lies in the fact that static images were taken into account. However, emotion is a dynamically changing state, therefore future work will concentrate on extending these techniques to video signals. Finally, the inherently unstable nature of adversarial learning can present various challenges related to finding the optimal number of iterations during the optimization for achieve a close-to-optimal solution. We used techniques like early stopping and random initialization of layers to tackle this, but more experimentation in terms of such hyper-parameters is needed in order to obtain more robust results.

\vspace{-8pt}
\section{Conclusion}
\vspace{-9pt}
To the best of our knowledge, this work is the first attempt to design an adversarial learning framework for preserving users' anonymity in face-based emotion recognition. We first provide a method to evaluate the amount of identity-specific information available in convolutional bases learned for emotion recognition. We then designed an adversarial learning framework through an approximate procedure which reduces the amount of user-dependent information in the convolutional layer, which is iteratively learned for various user-specific fully connected layers in an attempting to yield a robust emotion-specific and anonymity-preserving transformation. As part of our future work, we will expand this study to larger datasets and explore additional methods, such as adversarial reinforcement learning and self-attention mechanisms.

\bibliographystyle{IEEEbib}
\bibliography{strings,refs}

\end{document}